\documentclass[conference]{IEEEtran}
\IEEEoverridecommandlockouts
\usepackage{cite}
\usepackage{amsmath,amssymb,amsfonts}
\usepackage{algorithmic}
\usepackage{graphicx}
\usepackage{textcomp}
\usepackage{xcolor}
\usepackage[labelfont=bf]{caption}
\usepackage{hyperref}
\usepackage{amsmath}
\usepackage{multirow}

\usepackage[greek,latin,english]{babel}

\def\BibTeX{{\rm B\kern-.05em{\sc i\kern-.025em b}\kern-.08em
    T\kern-.1667em\lower.7ex\hbox{E}\kern-.125emX}}
\begin{document}

\title{A Grey-box Text Attack Framework using Explainable AI\\
\thanks{$*$Equal contribution. Listing order is alphabetical order. E. Chiramal and K. Soh Boon Kai, work are done during our stint at Amaris.AI Pte Ltd a AI-Cybersecurity Company based in Singapore}
}

\author{\IEEEauthorblockN{Esther Chiramal*}
Nanyang Technological University
\and
\IEEEauthorblockN{Kelvin Soh Boon Kai*}
University of Glasgow

}

\maketitle

\begin{abstract}
Explainable AI is a strong strategy implemented to understand complex black-box model predictions in a human interpretable language. It provides the evidence required to execute the use of trustworthy and reliable AI systems. On the other hand, however, it also opens the door to locating possible vulnerabilities in an AI model. Traditional adversarial text attack uses word substitution, data augmentation techniques and gradient-based attacks on powerful pre-trained Bidirectional Encoder Representations from Transformers (BERT) variants to generate adversarial sentences. These attacks are generally white-box in nature and not practical as they can be easily detected by humans E.g. Changing the word from “Poor” to “Rich”. We proposed a simple yet effective Grey-box cum Black-box approach that does not require the knowledge of the model while using a set of surrogate Transformer/BERT models to perform the attack using Explainable AI techniques. As Transformers are the current state-of-the-art models for almost all Natural Language Processing (NLP) tasks, an attack generated from BERT1 is transferable to BERT2. This transferability is made possible due to the attention mechanism in the transformer that allows the model to capture long-range dependencies in a sequence. Using the power of BERT generalisation via attention, we attempt to exploit how transformers learn by attacking a few surrogate transformer variants which are all based on a different architecture. We demonstrate that this approach is highly effective to generate semantically good sentences by changing as little as one word that is not detectable by humans while still fooling other BERT models. \\
\end{abstract}

\begin{IEEEkeywords}
BERT, Explainable AI, Grey-box Text Attack, Transformer
\end{IEEEkeywords}

\section{Introduction}
Adversarial attacks are an important and serious issue in the AI-Cybersecurity landscape. It opens doors for attackers to exploit the loopholes of AI models by misleading the AI model’s decision process. Due to the complexity of AI models it is difficult to determine if an adversarial attack has taken place leading to misclassified results. Adversarial attacks can fool facial recognition systems to approve authorised documents or modify images used in autonomous driving systems to cause accidents \cite{cite:fgsm}. Such misuses of AI can violate human values and cause detrimental impacts when used in industries having a direct connection to an individual’s lifestyle. The black-box nature of AI models pushes forth the requirement of having valid explanations for decisions made by AI systems. This is achieved through the concept of Explainable AI (XAI) that focuses on providing reasonings in a human interpretable form \cite{cite:xai,cite:xai_review}. Various explainable AI strategies that provide user-friendly visualisations have been implemented, mediating the issue of lack of transparency. The reasoning provided includes highlighting areas of a given data that contribute greatly to the final prediction made by AI systems. It is a powerful strategy used to instil trust in individual predictions and at the same time detect anomalies within the system. 
\\

Text classification, a machine learning technique, is one of the fundamental tasks in Natural Language Processing (NLP) and has various applications ranging from sentiment analysis to spam detection. The experiments in this paper are conducted on sentiment classifications incorporated with explainable AI strategies. As mentioned, Explainable AI is a powerful tool required for any decision making AI system. However, by opening the ‘black-box’ nature of AI systems and providing explanations, it reveals existing vulnerabilities within the system to all users. In this paper, we propose an algorithm to exploit the advantages of existing Explainable AI frameworks and construct an adversarial attack on AI models. The main contributions of the paper are summarised as follows.

\begin{itemize}
  \item Producing sentence variations by replacing words with synonyms of a given text based on information received through XAI strategies.
  \item Robustness of model and attack transferability between transformers.
\end{itemize}

\section{Related Work}

\subsection{TextAttack: A Framework for Adversarial Attacks, Data Augmentation, and Adversarial Training in NLP}
TextAttack is a framework to craft adversarial attacks in NLP. The framework is catered to work with various NLP models and Transformers that can be implemented through the use of traditional neural network and HuggingFace models \cite{cite:textattack}. TextAttack contains various algorithms to perform adversarial attack on classification tasks such as DeepWordBug, HotFlip, PWWS, TextBugger, and TextFooler. Most of the algorithms follow the idea of iterating through all possible combinations of potential transformations to find a sequence of successful alterations that can form adversarial samples. White-box and black-box transformations can be applied using TextAttack where white-box indicates that access to the model is available and black-box is conducted without any knowledge. White-box permutations allow attackers to determine words to change based on the model gradients; however, this is not possible for black-box transformations. 

\begin{figure}[htbp]
\centerline{\includegraphics[scale=0.2, width=93mm]{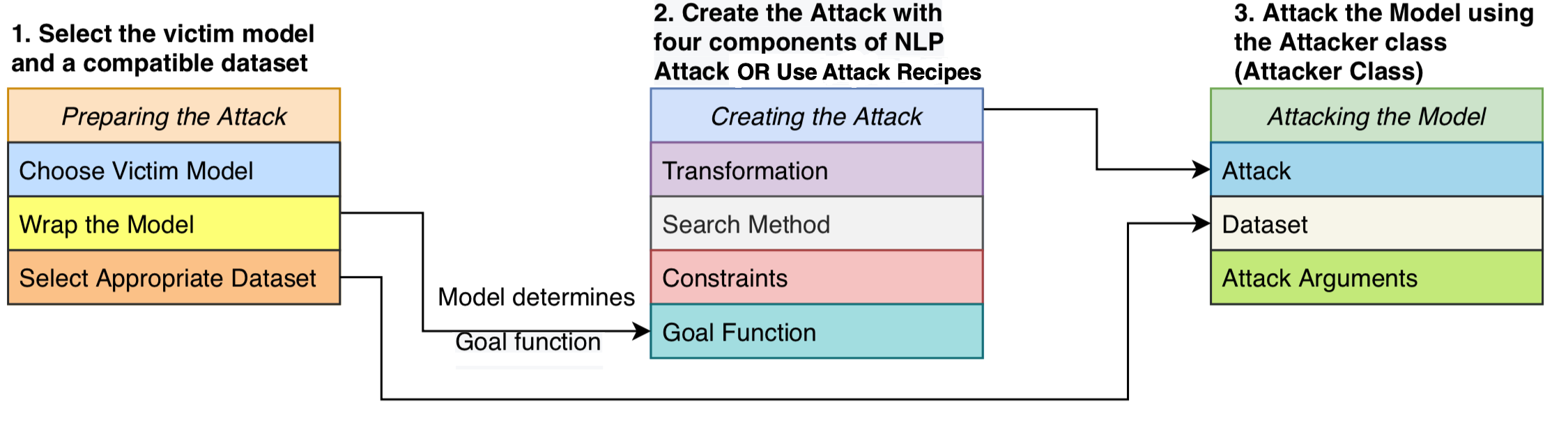}}
\caption{TextAttack framework for adversarial attacks, data augmentation, and model training in NLP}
\label{fig_text_attack}
\end{figure}

To conduct a fair comparison, we tested the TextAttack framework using a transformer model from HuggingFace pre-trained on the ‘IMDB’ dataset. Probability Weighted Word Saliency (PWWS) attack was implemented on the test samples since it generates sentences less likely to be detected by humans. PWWS is word substitution order introduced in 2019 \cite{cite:pwws} which aims to generate adversarial examples that maintain lexical correctness, grammatical correctness and semantic similarity. It is a technique where words are swapped with their synonyms based on their saliency score and maximum swap effectiveness. \\
\begin{table}[htbp]
\centering
\begin{tabular}{|c|c|}
\hline
\textbf{Original Text} & \textbf{Adversarial Text} \\ \hline
{\color[HTML]{000000} \begin{tabular}[c]{@{}c@{}}throws in enough clever \\ and unexpected twists \\ to \textcolor{red}{make} the formula\\ feel fresh\end{tabular}} & \begin{tabular}[c]{@{}c@{}}throws in enough clever\\ and unexpected twists to\\ \textcolor{red}{crap} the formula feel fresh\end{tabular} \\ \hline
\begin{tabular}[c]{@{}c@{}}\textcolor{red}{possibility} of bankruptcy. \\ \textcolor{red}{lack} of \textcolor{red}{assurance.}\\  Poor stability.\end{tabular} & \begin{tabular}[c]{@{}c@{}}theory of bankruptcy. \\ \textcolor{red}{want} of \textcolor{red}{sureness}. \\ Poor stability.\end{tabular} \\ \hline
\begin{tabular}[c]{@{}c@{}}the story gives ample \\ opportunity for large \\ scale action \\ and suspense, which\\  director shekhar kapur \\ supplies with \textcolor{red}{tremendous} skill\end{tabular} & \begin{tabular}[c]{@{}c@{}}the story gives ample \\ opportunity for large \\ scale action \\ and suspense, which\\  director shekhar kapur \\ supplies with \textcolor{red}{terrible} skill\end{tabular} \\ \hline
\end{tabular}
\caption{Comparison of original samples with their adversarial substitutions using PWWS. }
\label{table:tab1}
\end{table}

As shown in Table \ref{table:tab1}, successful adversarial samples are generated after swapping one or more words in the sentence. The saliency score tells us the contribution of the original word to the final prediction. The word is replaced if its contribution changes depending on its word saliency.

\subsection{Grey-Box Adversarial Attack and Defence for Sentiment Classification}
Grey-box text attack is not a well studied area, as performing an attack with limited to no knowledge of the model can be challenging. A paper was released by researchers at IBM that proposes a framework to produce high quality adversarial samples while simultaneously training a classifier for adversarial defence \cite{cite:gradientmethods}. The architecture makes use of a generator which can be implemented as an autoencoder or a paraphrase generator. This generator is fed into two copies of a pre-trained classifier where one copy is updated during the training process with adversarial samples while the second copy is used to predict labels of the adversarial texts. \\

\begin{figure}[htbp]
\centerline{\includegraphics[scale=0.35]{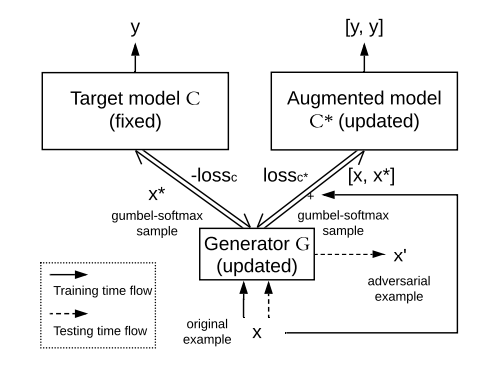}}
\caption{Framework architecture from \textit{“Grey-box Adversarial Attack and Defence for Sentiment Classification”}}
\label{fig_semattack}
\end{figure}

The framework however, requires details regarding the model architecture and weights apriori and hence cannot be directly applied to other classifier models. As shown in Figure \ref{fig_semattack}, the results of the adversarial attack are obtained through a joint architecture between the generator $G$ and the target model $C$. The paper assumes complete access to the target model during the training process which is not applicable for real-world applications. It is difficult to obtain details regarding model weights that are used in objective functions to generate adversarial samples.

\subsection{LIME: Local Interpretable Model-Agnostic Explanations}
LIME is an algorithm that allows any classifier or regressor to explain their decision outcome by approximating it locally with an interpretable model. It is an open source package created by Marco Tulio Ribeiro, Sameer Singh and Carlos Guestrin which focuses on explaining what machine learning classifiers are doing \cite{cite:lime}. To understand how the model arrive at a decision, we required a representation that is understandable to humans, regardless of the features that were used for training the model. For example, in the case of text classification a classifier may use complex features such as word embeddings. However, with the help of binary vectors this representation can be simplified into a more interpretable form. \\

\begin{figure}[htbp]
\centerline{\includegraphics[scale=0.3]{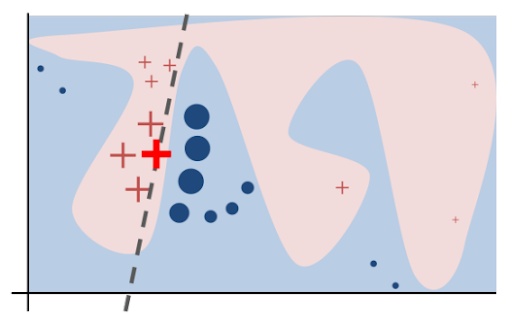}}
\caption{Toy example of the concept behind LIME (Image from “Why Should I Trust You?” Explaining the Predictions of Any Classifier” \cite{cite:xai})}
\label{fig_lime}
\end{figure}

The primary institution behind LIME is presented in below Figure \ref{fig_lime} where instances are sampled both in the vicinity of a chosen point, represented by a red cross, and also instances far away from the point. LIME samples instances around $X$, gets the predictions and weighs them by the proximity to the instance being explained. Local surrogate models are trained on these weighted instances that produce an interpretable model used to determine the class of the text of interest. It is a strong Explainable AI strategy which can be used to mitigate the issue of poor understanding of black-box models.

Many algorithms in the TextAttack framework including PWWS use white-box techniques which require access to the model parameters to calculate the gradients. However,  in the real world it is almost impossible to access the target model and the underlying datasets used for training the target model.  Gong et al., \cite{cite:gradientmethods} uses word vector spaces such as GloVe to search for adversarial embeddings using gradient methods in the embedding space in order to generate adversarial texts. Ren et al., \cite{cite:pwws} used PWWS that considers the word saliency as well as the classification probability. Both methods generate high-quality adversarial texts, however their framework requires the calculation of model gradients which is not feasible for practical and unknown models. \\

Our work extends the two ideas mentioned above by modifying the requirement of model gradients to determine saliency scores with Explainable AI which can be implemented without in-depth knowledge of the target models (Grey-box). Furthermore, it can be used as a robust testing framework for measuring the strength and transferability of adversaries between Transformers.

\section{Design}
Our framework is a combination of two main parts. \textbf{1. Generation of adversarial examples via exploitation of XAI, 2. Validation of Model Robustness and Attack Transferability through Black-box testing}. Both approaches do not require any prior knowledge of the target model allowing an effective attack and testing to be done. The following section A – E will describe the overview of our Grey-box Text Attack Framework, with an emphasis on the Transformer.

\subsection{BERT \& Transformer}\label{transformer}
Bidirectional Encoder Representations from Transformers, also known as BERT, is an open-source machine learning framework for NLP. The basic architecture of BERT is based on the Transformer architecture by stacking twelve encoder blocks without the decoder blocks as compared to Transformer. We will be using BERT and Transformer interchangeably \cite{cite:bert, cite:survey_bert}.  \\

Transformers are the current state-of-the-art models used for NLP tasks. They are large language models that are pre-trained on large corpuses such as Common Crawl to provide generalisation to downstream tasks \cite{cite:attention}. Our design is tested on various transformer models that have been pre-trained and can be obtained from HuggingFace \cite{cite:huggingface}. These pre-trained models are useful since they are well suited for many downstream tasks, which we will be using in our experiments.  The generalisation ability of transformers holds for different architectures, however, the better the generalisation, the better the transferability of adversarial attacks between transformers which we will see later.  \\

Furthermore, the attention mechanism in transformers has revolutionised the NLP landscape, by processing sentences as a whole achieving state-of-the-art performance for many NLP tasks. These attention modules in the transformer allow each token in the sequence to attend to every other token in the sequence, providing context and order of the words for any position in the sequence via positional encoding \cite{cite:attention}.

\begin{figure}[htbp]
\centerline{\includegraphics[scale=0.2]{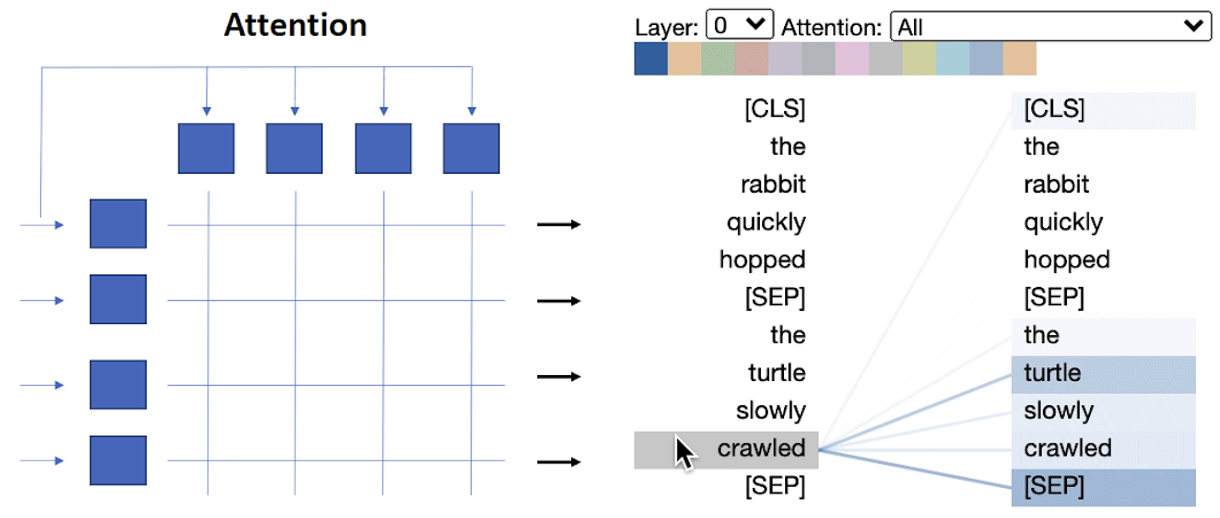}}
\caption{Self-Attention compares all input sequence tokens (Square blocks) with each other allowing them to understand long range dependencies between tokens.}
\label{fig_self_attention}
\end{figure}

\subsection{LIME used to obtain word contributions}
The concept of Explainable AI is to provide satisfactory explanations for predictions made by AI systems for its expected input, output, and potential biases. It assists in the validation of  model accuracy, fairness, transparency and outcomes of AI-powered decision making which is crucial for building trust and confidence in the systems. To our limited knowledge, no work has been done in using XAI to exploit the vulnerability of transformer models.  \\

\begin{figure}[htbp]
\centerline{\includegraphics[scale=0.4, width=90mm]{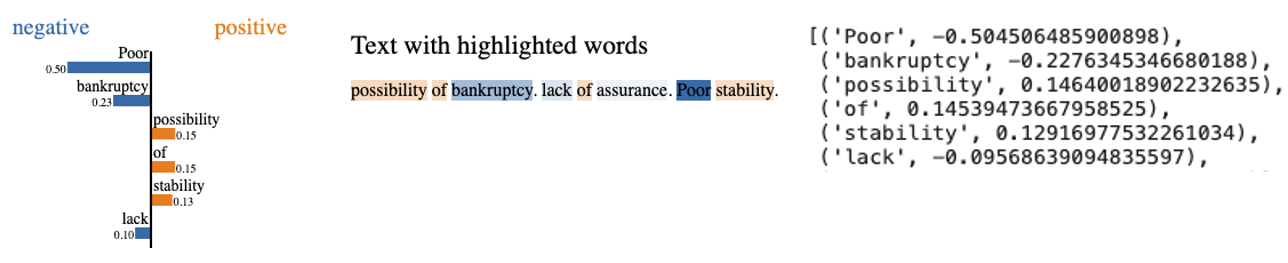}}
\caption{LIME output for a given text along with a list displaying word contributions.}
\label{fig_lime_output_lime_score}
\end{figure}

Using XAI allows us to quickly determine the word contribution of every token in a sentence that leads to the final prediction of a model. Since every token in a sentence carries some information that eventually leads to the final model prediction, changing a word with the highest contribution should have a greater effect at manipulating the outcome of the target model prediction than a word with a lower contribution. Figure \ref{fig_lime_output_lime_score} shows the LIME output obtained for a text passed through a pre-trained Transformer model. The output provides a list of words in the given text along with their contribution values. This information is valuable as it can be used to identify the words that have to be modified in order of their importance to change the predicted label. 

\clearpage

From the example shown in Figure \ref{fig_lime_output_lime_score}, it can be identified that changing the words \textbf{‘poor’} or \textbf{‘possibility’} can have a greater effect on the contribution values and increase the chances of fooling a model. Therefore, the use of XAI presents an opportunity to identify vulnerabilities in the used transformer model which can then be exploited to produce successful adversarial attacks.

\subsection{Synonym Substitution}
In order to ensure that the adversarial sentences generated are similar to the original text, it is important to refrain from adding additional words, corrupting the sentences and changing the sentence semantics by replacing words with their antonyms. All these will have a negative impact on the final sentence as it can be easily detected by humans. The idea we propose is similar to PWWS where words are replaced with their closest synonyms based on the WordNet corpus provided by NLTK (Natural Language Toolkit). WordNet is  a large lexical database of  English language to produce synonyms for the words found using the LIME explainer. These synonyms are then used to replace the original word in the sentence producing a list of new sentences. For example, replacing “\textbf{Hi}” with “\textbf{Hello}” or “\textbf{Average}” with “\textbf{average}”, making subtle changes to sentences that is enough to fool both humans and models.  \\

\begin{figure}[htbp]
\centerline{\includegraphics[scale=0.4, width=90mm]{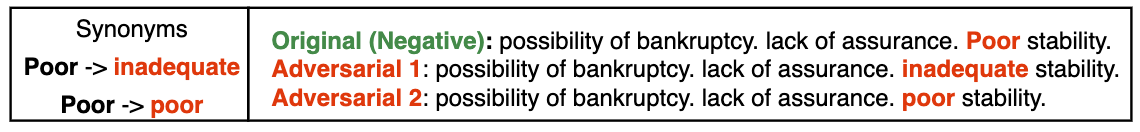}}
\caption{Synonym Substitution using one word.}
\label{fig_syn_sub_one_word}
\end{figure}

A greedy attack is implemented where every sentence containing a modification with a synonym is tested against the target model. The algorithm begins with the modification of 1 word followed by N words depending on the minimum number of words required to be changed in order for the final prediction to be reversed. Examples of sentences generated through synonyms swaps are shown below Figure \ref{fig_syn_sub_one_word}.

\subsection{Surrogate Model}
\label{surrogate_model}
Surrogate models are used to produce a grey-box framework that ensures the transferability of the attack by testing generated samples against these models before applying them on the target model.  This is to achieve successful attacks with little or no knowledge of the target models. Given N combinations of sentences generated from the previous synonym substitution, we can freely pick any transformer model to test the transferability of the attack. The idea of picking a transformer model serves as a surrogate model to validate the effectiveness of adversarial transferability via XAI and synonyms substitution. This serves as a feedback mechanism where we are able to repeatedly query transformer models in an attempt to fool them by probing for a different label. The following Figure \ref{fig_proposed_framework} illustrates our proposed framework. *\textbf{Note: All tested Transformer model are based on the base variant as compared to the large variant 12 encoder layers vs 24 encoder layers}.

\begin{figure}[htbp]
\centerline{\includegraphics[scale=0.32]{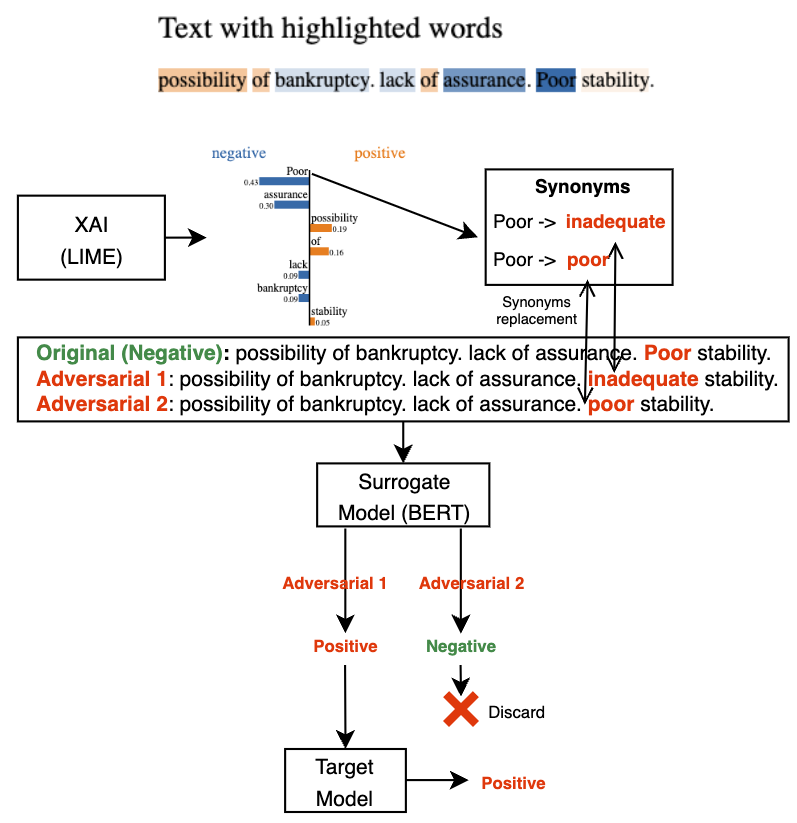}}
\caption{Proposed Grey-box Text Attack Framework using Explainable AI}
\label{fig_proposed_framework}
\end{figure}

\subsection{Scaling up Surrogate Model}
\begin{figure}[htbp]
\centerline{\includegraphics[scale=0.25]{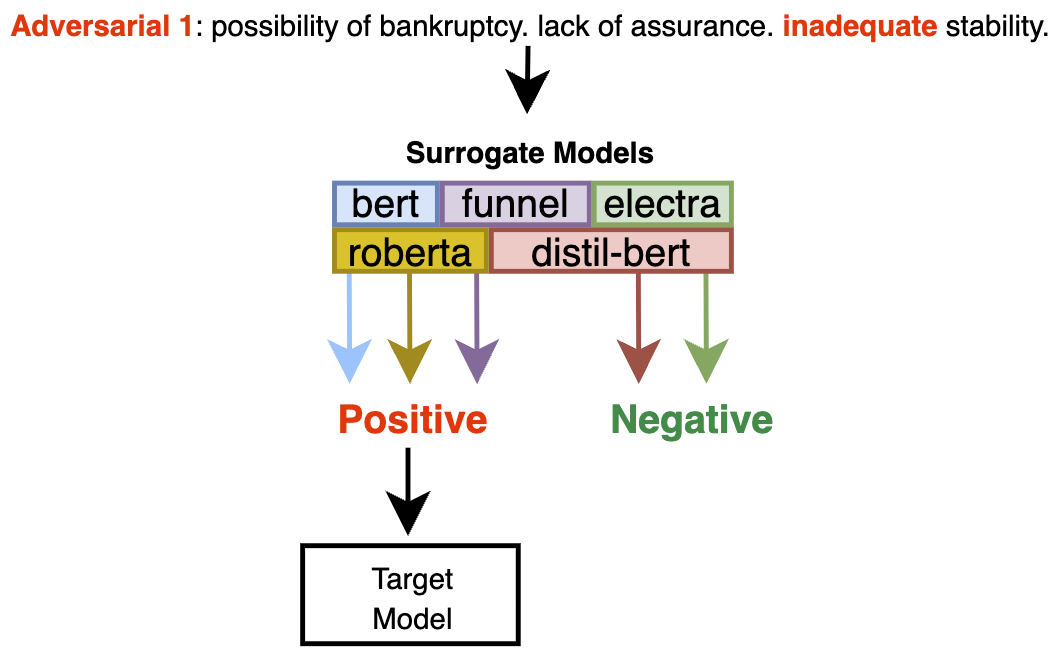}}
\caption{High level overview of surrogate model testing based on a generated adversarial sample. The larger the number of surrogate models fooled the higher the possibility of the given instance to transfer the attack across other model architectures.}
\label{fig_proposed_framework_n_test}
\end{figure}

Instead of using only one surrogate model to validate the adversarial sample, we can scale this up to $N$ surrogate models all based on a different architecture, let $N_s$ denote the number of surrogates model that can be fooled by the adversarial sample. This idea allows all the surrogate models to find an adversarial sample that can fool not only one but $N$ surrogate models if and only if they all agree on the same adversarial sample prediction displaying a domino effect using a predetermined threshold of $N_s \geq \lceil \frac{N}{2} \rceil$. The testing of multiple surrogate models was conducted on adversarial sentences generated based on a given sentence \textit{“possibility of bankruptcy. lack of assurance. Poor stability”}. Figure \ref{fig_proposed_framework_n_test} shows an instance of testing where adversarial samples are passed through multiple BERT variations and the number of successful model attacks are recorded. In-depth details regarding the testing results can be found in Table \ref{table:tab2} in Section \ref{strength_of_surrogate_models}.

\section{Evaluation} \label{evaluation}

\subsection{Dataset}

\subsubsection{IMDB}
Our framework was tested on the IMDB movie reviews dataset. The dataset contains 25,000 highly polar movie reviews for training and another 25,000 reviews for testing. The IMDB dataset is widely adapted for binary classification applications for text. To date there are more than 474 Transformers trained on IMDB datasets in HuggingFace. \\

\subsection{Model}

All the models used in the experiment were from HuggingFace Transformer \cite{cite:huggingface} which allows us to conduct rapid testing and experiments on many different architectures.

\subsection{Strength of Surrogate Models against Adversarial Sentences}\label{strength_of_surrogate_models}
To validate the effectiveness of adversarial sentences that can fool other transformer models, we experiment on a sentence \textit{“possibility of bankruptcy. lack of assurance. Poor stability.”} with a \textit{“negative”} label that generates 12 adversarial sentences $N_{sent}$ after a single synonym substitution. These 12 adversarial sentences are tested on different transformers to illustrate the strength of the adversarial sentences at fooling the different models. We use different transformer architectures for the purpose of diversity. (Please refer to the appendix for the generated adversarial sentences.) Table \ref{table:tab2} below highlights the percentage of sentences that successfully attacked a given model along with an average confidence score of the predicted label across all successful sentences.

\newpage
\begin{equation} \label{eq:test}
\begin{gathered}
\text{\% of sentences} = \frac{N_{succ}}{N_{sent}} \times 100 \\
\text{Average Confidence} = \frac{1}{N_{succ}}\sum_{i=0}^{N_{succ}} \text{(label confidence)} \\
\end{gathered}
\end{equation}

where $N_{sent}$ = Number of sentences, $N_{succ}$ = Number of successful sentences,  and (label confidence) = Percentage contribution to predicted class label.

\begin{table}[!htbp]
\centering
\begin{tabular}{|c|c|c|}
\hline
\textbf{Surrogate model} & \textbf{\begin{tabular}[c]{@{}c@{}}\% of sentences that \\ successfully  fool surrogate \\ models  ($N_{succ}$ out \\ of $N...N_{sent}$* \\ generated  sentences)\end{tabular}} & \textbf{\begin{tabular}[c]{@{}c@{}}Average \% confidence\\  for all adversarial \\ sentences\\  (Predicted label \\ = Positive)\end{tabular}} \\ \hline
\textbf{BERT} & 91\% & 81\% \\ \hline
\textbf{RoBERTa-base} & 8.3\% & 91\% \\ \hline
DistilBERT & - & - \\ \hline
\textbf{Funnel-base} & 33.3\% & 63.4\% \\ \hline
\end{tabular}
\caption{Evaluation of different transformer models used against adversarial test samples generated using our proposed framework. \textbf{*For the experiment conducted using $N_{sent} = 12$ adversarial sentences}}
\label{table:tab2}
\end{table}

From the evaluation conducted above, based on the specific example used, BERT base and RoBERTa models were successfully fooled by some of the sample adversarial generated. The samples however failed to manipulate predictions of the DistilBERT model.

\subsection{Between Surrogates Transferability} \label{between_surrogates_transferability}
Adversarial sentences generated after synonyms substitutions are tested against the different Transformer variants that acts as a surrogate to obtain a list of adversarial samples that manages to fool at least 2 out of the 4 models. Table \ref{table:tab3} below shows the results obtained for three sentences that were tested on various Transformer models.  From Table \ref{table:tab3}, it can be inferred that the BERT and Electra models were fooled by the adversarial samples. In the given experiment, the adversarial sentence \textit{‘possibility of bankruptcy. lack of assurance. \textbf{short} stability.’} manages to attack at least 2 of the models used for the testing. This shows that an adversarial sentence generated from one BERT model is transferable across different architectures.

\begin{table}[!htbp]
\centering
\begin{tabular}{c|ccc|}
\cline{2-4}
 & \multicolumn{3}{c|}{\textbf{Surrogate Model}} \\ \hline
\multicolumn{1}{|c|}{\textbf{Adversarial Text}} & \multicolumn{1}{c|}{\textbf{BERT}} & \multicolumn{1}{c|}{\textbf{DistilBERT}} & \textbf{ELECTRA} \\ \hline
\multicolumn{1}{|c|}{\begin{tabular}[c]{@{}c@{}}possibility of bankruptcy. \\ lack of assurance.\\  \textcolor{red}{hapless} stability.\end{tabular}} & \multicolumn{1}{c|}{\begin{tabular}[c]{@{}c@{}}Positive \\ 89.7\%\end{tabular}} & \multicolumn{1}{c|}{\textbf{x}} & \begin{tabular}[c]{@{}c@{}}Positive\\ 52.7\%\end{tabular} \\ \hline
\multicolumn{1}{|c|}{\begin{tabular}[c]{@{}c@{}}possibility of bankruptcy. \\ lack of assurance. \\ \textcolor{red}{short} stability.\end{tabular}} & \multicolumn{1}{c|}{\begin{tabular}[c]{@{}c@{}}Positive \\ 74.8\%\end{tabular}} & \multicolumn{1}{c|}{\textbf{x}} & \begin{tabular}[c]{@{}c@{}}Positive \\ 50.8\%\end{tabular} \\ \hline
\multicolumn{1}{|c|}{\begin{tabular}[c]{@{}c@{}}possibility of bankruptcy. \\ lack of assurance. \\ \textcolor{red}{inadequate} stability.\end{tabular}} & \multicolumn{1}{c|}{\begin{tabular}[c]{@{}c@{}}Positive \\ 80.1\%\end{tabular}} & \multicolumn{1}{c|}{\textbf{x}} & \begin{tabular}[c]{@{}c@{}}Positive \\ 50.3\%\end{tabular} \\ \hline
\end{tabular}
\caption{Table depicting the results for surrogate model testing. (\textbf{x} denotes that the model is not fooled (no change in label))}
\label{table:tab3}
\end{table}

\subsection{Adversarial Attack Transferability} 
To quantify the success rate of an adversarial sentence, we use the number of surrogate models as a proxy to gauge the likelihood of a successful surrogate attack. This was done with a $\lceil \frac{N}{2} \rceil$ threshold where N denotes the number of surrogates used for testing and fooling the surrogates model with a successful attack. An effective transfer rate based on empirical testing is $N_s \geq \lceil \frac{N}{2} \rceil$ as shown in Section \ref{between_surrogates_transferability}. The transferability rate dramatically improves when scaling up to more surrogates. As shown in the below Table \ref{table:tab4}, we tested on three target models of different architectures that were not part of the surrogates testing (BERT, DistilBERT and ELECTRA) which allowed us to transfer the attack successfully from \textit{“Negative”} to \textit{“Positive”}. All target models predicted the original sentence with a \textit{“Negative”} label of \> 90\% confidence. \\

\textbf{Original sentence}: possibility of bankruptcy. lack of assurance. \textbf{Poor} stability. $\Rightarrow$ \textit{Negative} \\
\textbf{Adversarial sentence}: possibility of bankruptcy. lack of assurance. \textcolor{red}{\textbf{short}} stability. \\

\begin{table}[!htbp]
\centering
\begin{tabular}{|c|c|c|}
\hline
\textbf{\begin{tabular}[c]{@{}c@{}}Target \\ model\end{tabular}} & \textbf{\begin{tabular}[c]{@{}c@{}}\% confidence for a \\ adversarial \\ sentence that fooled \\ 2 out of 3 \\ surrogate models \\ (BERT, DistilBERT, \\ ELECTRA)\end{tabular}} & \textbf{\begin{tabular}[c]{@{}c@{}}Predicted Label for\\  Adversarial \\ sentence\end{tabular}} \\ \hline
\textbf{XLM-RoBERTa} & 50.2\% & \textbf{Positive} \\ \hline
\textbf{RoBERTa} & 91.4\% & \textbf{Positive} \\ \hline
\textbf{Funnel} & 67.3\% & \textbf{Positive} \\ \hline
\end{tabular}
\caption{Confidence value of predictions tested on the above adversarial sentence. }
\label{table:tab4}
\end{table}

\section{Results}
The table below shows instances of successful adversarial attacks after the implementation of the framework. These sentences have managed to fool 2 or more surrogate models and are applied on a target model with a different architecture or trained dataset (similar sentiment).

\begin{table}[htbp]
\centering
\begin{tabular}{|c|c|c|}
\hline
\begin{tabular}[c]{@{}c@{}}\textbf{Original}\\ \textbf{Prediction}\end{tabular} & \begin{tabular}[c]{@{}c@{}}\textbf{Adversarial}\\ \textbf{Prediction}\end{tabular} & \textbf{Adversarial Sentence} \\ \hline
\begin{tabular}[c]{@{}c@{}}Negative\\ 98.6\%\end{tabular} & \begin{tabular}[c]{@{}c@{}}Positive\\ 80.1\%\end{tabular} & \begin{tabular}[c]{@{}c@{}}possibility of bankruptcy. lack of \\ assurance. (Poor -\textgreater \textcolor{red}{Inadequate})\\  stability.\end{tabular} \\ \hline
\begin{tabular}[c]{@{}c@{}}Negative\\ 96.2\%\end{tabular} & \begin{tabular}[c]{@{}c@{}}Positive\\ 98.2\%\end{tabular} & \begin{tabular}[c]{@{}c@{}}Applicant has a (poor -\textgreater \textcolor{red}{short}) \\ credit history.\\  It will be hard to repay loans. \\ Average investments and insurance \\ coverage. \\ Requires time to pay debts.\end{tabular} \\ \hline
\begin{tabular}[c]{@{}c@{}}Negative\\ 99.9\%\end{tabular} & \begin{tabular}[c]{@{}c@{}}Positive\\ 74.3\%\end{tabular} & \begin{tabular}[c]{@{}c@{}}All in all, every character in this film \\ is somebody that very few  ...  \\ it's an irritating snore fest. \\ That's what happens when you're \\ out of touch. You entertain your\\  few friends with inside jokes, \\ and (bore -\textgreater \textcolor{red}{yield}) all the rest.\end{tabular} \\ \hline
\end{tabular}
\caption{Successful adversarial samples.}
\label{table:tab6}
\end{table}

From the results shown in the Table \ref{table:tab6} above, it can be inferred that the models can be fooled with minor changes to the text that can go undetected during human monitoring. Depending on the model being tested, the reliability of the results varies. This is because pre-trained models can have different performance levels based on their specified hyperparameters. Hence, the use of surrogate models as mentioned in Section \ref{surrogate_model} is important to obtain adversarial samples that are able to attack multiple models ensuring a higher probability of transferability of attacks.

\section{Further Area of Research}

\subsection{Post-processing of Adversarial sentences}
In a situation, where the adversarial sentences fail to attack a given model based on synonym swaps, additional NLP techniques can be implemented to further modify the sentence. One of these techniques is the insertion of unicode characters into sentences to complement the synonym swaps. Such attacks have been discussed by Nicholas Boucher et al. \cite{cite:unicodeattack} where unicodes can be used to insert invisible characters or homoglyphs, or modify and delete special characters from the sentences. Table \ref{tab6} shows a simple attack conducted by replacing all instances of ‘i’ in a sentence with its homoglyphical replacement, latin small letter ‘\textlatin{i}’.

\begin{table}[htbp]
\centering
\begin{tabular}{|c|c|c|}
\hline
\begin{tabular}[c]{@{}c@{}}\textbf{Original} \\ \textbf{Prediction}\end{tabular} & \begin{tabular}[c]{@{}c@{}}\textbf{Adversarial} \\ \textbf{Prediction}\end{tabular} & \begin{tabular}[c]{@{}c@{}}\textbf{Adversarial}\\  \textbf{Sentence}\end{tabular} \\ \hline
\begin{tabular}[c]{@{}c@{}}Negative\\ 99.7\%\end{tabular} & \begin{tabular}[c]{@{}c@{}}Positive\\ 84.8\%\end{tabular} & \begin{tabular}[c]{@{}c@{}}poss\textcolor{red}{\textlatin{i}}b\textcolor{red}{\textlatin{i}}l\textcolor{red}{\textlatin{i}}ty of bankruptcy. \\ lack of assurance. Inadequate \\ stab\textcolor{red}{\textlatin{i}}l\textcolor{red}{\textlatin{i}}ty.\end{tabular} \\ \hline
\end{tabular}
\caption{Example of a possible unicode attack. (Highlighted \textcolor{red}{\textlatin{i}} denotes the latin small letter ‘\textlatin{i}’)}
\label{tab6}
\end{table}

A combination of synonym and unicode attack increases the robustness and reliability of the sentence in manipulating various types of models. 

\section{Conclusion}
Our Grey-box Text Attack Framework using Explainable AI is a simple yet effective technique used to generate higher quality and better texts by modifying a minimum number of words and maintaining the existing sentiment of the text. We emphasise on the use of Explainable AI to provide information regarding word contributions in order to obtain vulnerable instances in a given model to frame a possible attack. In this framework, we use XAI to derive a sorted list of words in the order of highest contribution to generate better adversarial sentences by swapping the words with their synonyms. Moreover, the framework allows the testing of the adversarial samples against multiple surrogate models that can be customised according to a user’s preference. This idea can be further extended with the use of other explainable AI strategies not limited to LIME. \\

Furthermore we do not require deep knowledge of the model such as the architecture and gradients of the model. However minor details regarding type of classification i.e sentiment, binary or multiclass is required along with the class labels in order to implement LIME and other XAI techniques. In the future, we would like to enhance our attacking strategies by implementing more NLP attacks inclusive of unicode variations and experiment with longer sentences and stronger models that are adversarially trained. Finally, we can also extend this framework to validating model robustness against XAI exploitation and synonyms substitutions that are data and model agnostic.

\section*{Appendix}
In the Appendix, we add the list of adversarial sentences generated that were used for testing

\clearpage
\onecolumn

\begin{table}[!htbp]
\centering
\begin{tabular}{|c|c|c|}
\hline
\textbf{Original Sentence} & \textbf{Adversarial Sentence} & \multicolumn{1}{c|}{\begin{tabular}[c]{@{}c@{}}No. of\\ Surrogate Models\\ Fooled\end{tabular}} \\ \hline
\multirow{8}{*}{possibility of bankruptcy. lack of assurance.  \textbf{Poor} stability.} & possibility of bankruptcy. lack of assurance. inadequate stability. & 1 \\ \cline{2-3} 
 & possibility of bankruptcy. lack of assurance. \textcolor{red}{pitiable} stability & 1 \\ \cline{2-3} 
 & possibility of bankruptcy. lack of assurance.  \textcolor{red}{poor} stability. & 1 \\ \cline{2-3} 
 & possibility of bankruptcy. lack of assurance. \textcolor{red}{piteous} stability. & 2 \\ \cline{2-3} 
 & possibility of bankruptcy. lack of assurance. \textcolor{red}{miserable} stability. & 3 \\ \cline{2-3} 
 & possibility of bankruptcy. lack of assurance. \textcolor{red}{hapless} stability. & 2 \\ \cline{2-3} 
 & possibility of bankruptcy. lack of assurance. \textcolor{red}{misfortunate} stability. & 3 \\ \cline{2-3} 
 & possibility of bankruptcy. lack of assurance. \textcolor{red}{short} stability. & 4 \\ \hline
\multirow{3}{*}{\begin{tabular}[c]{@{}c@{}}All in all, every character in this film is somebody \\ that very few people can relate with,\\ unless you're millionaire \\from Manhattan  with beautiful supermodels \\ at your beckon call. \\For the rest of us, it's an irritating snore fest. \\That's what happens when you're out of touch. \\ You entertain your few friends with \\inside jokes, and \textbf{bore} all the rest\end{tabular}} & \begin{tabular}[c]{@{}c@{}}All in all, every character in this film is somebody \\ that very few people can relate with,\\ unless you're millionaire \\from Manhattan  with beautiful supermodels \\ at your beckon call. \\For the rest of us, it's an irritating snore fest. \\That's what happens when you're out of touch. \\ You entertain your few friends with \\inside jokes, and  \textcolor{red}{suffer} all the rest\end{tabular} & 1 \\ \cline{2-3} 
 & \begin{tabular}[c]{@{}c@{}}All in all, every character in this film is somebody \\ that very few people can relate with,\\ unless you're millionaire \\from Manhattan  with beautiful supermodels \\ at your beckon call. \\For the rest of us, it's an irritating snore fest. \\That's what happens when you're out of touch. \\ You entertain your few friends with \\inside jokes, and  \textcolor{red}{endure} all the rest\end{tabular} & 1 \\ \cline{2-3} 
 & \begin{tabular}[c]{@{}c@{}}All in all, every character in this film is somebody \\ that very few people can relate with,\\ unless you're millionaire \\from Manhattan  with beautiful supermodels \\ at your beckon call. \\For the rest of us, it's an irritating snore fest. \\That's what happens when you're out of touch. \\ You entertain your few friends with \\inside jokes, and  \textcolor{red}{yield} all the rest.\end{tabular} & 2 \\ \hline
\end{tabular}
\caption{List of adversarial samples used for testing and fooling different surrogate models. (\textcolor{red}{Red} highlighted word denotes the synonym of the replaced \textbf{bold} word in the Original Text)}
\label{table:tab7}
\end{table}

\end{document}